\documentclass[times,twocolumn,final]{elsarticle}

\usepackage{framed,multirow}

\usepackage{amsmath}
\usepackage{amssymb}
\usepackage{bm}

\usepackage{pdflscape}
\usepackage{afterpage}

\usepackage[group-separator={,}]{siunitx}

\usepackage{url}
\usepackage{xcolor}

\usepackage{hyperref}

\usepackage[figuresright]{rotating}

\definecolor{newcolor}{rgb}{.8,.349,.1}

\journal{arXiv}

\begin{document}


\begin{frontmatter}

\title{Quantifying the effects of data augmentation and stain color normalization in convolutional neural networks for computational pathology}%

\author[1]{David Tellez\corref{cor1}}
\cortext[cor1]{Corresponding author.}
\ead{david.tellezmartin@radboudumc.nl}

\author[1]{Geert Litjens}
\author[1]{P\'eter B\'andi}
\author[1]{Wouter Bulten}
\author[1]{John-Melle Bokhorst}
\author[1]{Francesco Ciompi}
\author[1,2]{Jeroen van der Laak}

\address[1]{Diagnostic Image Analysis Group and the Department of Pathology, Radboud University Medical Center, Nijmegen, The Netherlands}
\address[2]{Center for Medical Image Science and Visualization, Link\"oping University, Link\"oping, Sweden}


\begin{abstract}
Stain variation is a phenomenon observed when distinct pathology laboratories stain tissue slides that exhibit similar but not identical color appearance. Due to this color shift between laboratories, convolutional neural networks (CNNs) trained with images from one lab often underperform on unseen images from the other lab. Several techniques have been proposed to reduce the generalization error, mainly grouped into two categories: stain color augmentation and stain color normalization. The former simulates a wide variety of realistic stain variations during training, producing stain-invariant CNNs. The latter aims to match training and test color distributions in order to reduce stain variation. For the first time, we compared some of these techniques and quantified their effect on CNN classification performance using a heterogeneous dataset of hematoxylin and eosin histopathology images from 4 organs and 9 pathology laboratories. Additionally, we propose a novel unsupervised method to perform stain color normalization using a neural network. Based on our experimental results, we provide practical guidelines on how to use stain color augmentation and stain color normalization in future computational pathology applications. 
\end{abstract}



\end{frontmatter}


\section{Introduction}

\begin{figure*}[h!]
\centering
\includegraphics[width=1.0\textwidth]{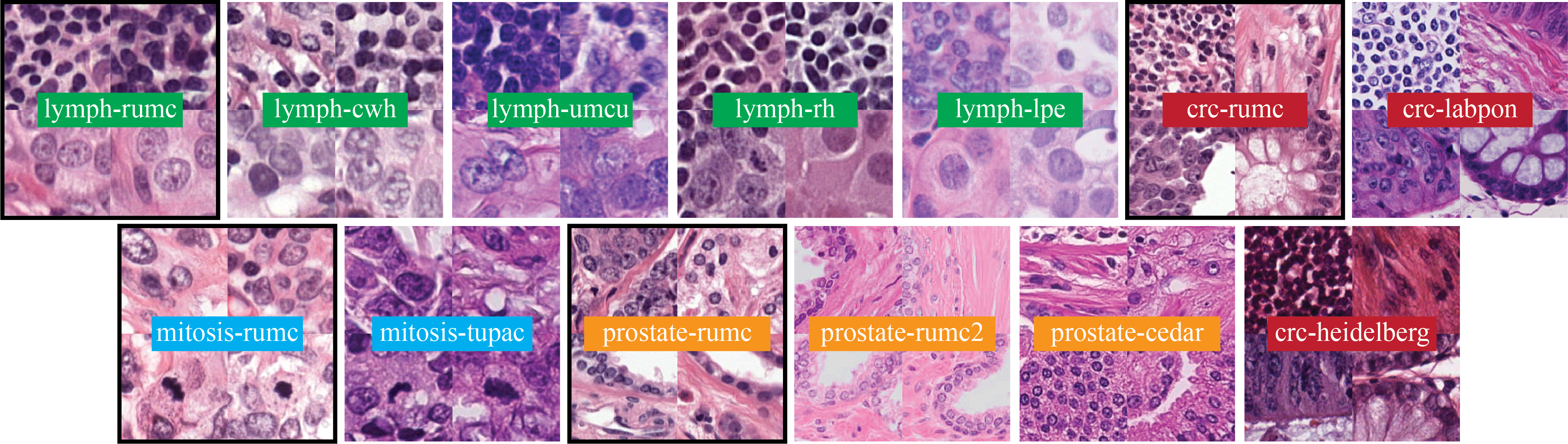}
\caption{\label{fig:patches} Example images from training and test datasets. Applications are indicated by colors and keywords: tumor detection in lymph nodes (\textit{lymph}), colorectal cancer tissue classification (\textit{crc}), mitosis detection (\textit{mitosis}) and prostate epithelium detection (\textit{prostate}). Training set images are indicated by the keyword \textit{rumc} and black outline. The rest belong to test sets from other centers. Stain variation can be observed between training and test images.}
\end{figure*}

Computational pathology aims at developing machine learning based tools to automate and streamline the analysis of whole-slide images (WSI), i.e. high-definition images of histological tissue sections. These sections consist of thin slices of tissue that are stained with different dyes so that tissue architecture becomes visible under the microscope. In this study, we focus on hematoxylin and eosin (H\&E), the most widely used staining worldwide. It highlights cell nuclei in blue color (hematoxylin), and cytoplasm, connective tissue and muscle in various shades of pink (eosin). The eventual color distribution of the WSI depends on multiple steps of the staining process, resulting in slightly different color distributions depending on the laboratory where the sections were processed, see Fig.~\ref{fig:patches} for examples of H\&E stain variation. This inter-center stain variation hampers the performance of machine learning algorithms used for automatic WSI analysis. Algorithms that were trained with images originated from a single pathology laboratory often underperform when applied to images from a different center, including state-of-the-art methods based on convolutional neural networks (CNNs)~(\cite{Goodfellow-et-al-2016,komura2018machine,veta2018predicting,sirinukunwattana2017gland}). Existing solutions to reduce the generalization error in this setting can be categorized into two groups: (1)~\textit{stain color augmentation}, and (2)~\textit{stain color normalization}.

\subsection{Stain color augmentation}

Stain color augmentation, and more generally data augmentation, has been proposed as a method to reduce CNN generalization error by simulating realistic variations of the training data. These artificial variations are hand-engineered to mimic the appearance of future test samples that deviate from the training manifold. Previous work on data augmentation for computational pathology has defined two main groups of augmentation techniques: (1) morphological and (2) color transformations~(\cite{liu2017detecting,tellez2018whole}). Morphological augmentation spans from simple techniques such as 90-degree rotations, vertical and horizontal mirroring, or image scaling; to more advanced methods like elastic deformation~(\cite{simard2003best}), additive Gaussian noise, and Gaussian blurring. The common denominator among these transformations is the fact that only the morphology of the underlying image is modified and not the color appearance, e.g. Gaussian blurring simulates out of focus artifacts which is a common issue encountered with WSI scanners. Conversely, color augmentation leaves morphological features intact and focuses on simulating stain color variations instead. Common color augmentation techniques borrowed from Computer Vision include brightness, contrast and hue perturbations. Recently, researchers have proposed other approaches more tailored to mimic specific H\&E stain variations, e.g. by perturbing the images directly in the H\&E color space~(\cite{tellez2018whole}), or perturbing the principal components of the pixel values~(\cite{bug2017context}). 

\subsection{Stain color normalization}

Stain color normalization reduces stain variation by matching the color distribution of the training and test images. Traditional approaches try to normalize the color space by estimating a color deconvolution matrix that allows identifying the underlying stains~(\cite{reinhard2001color,macenko2009method}). More recent methods use machine learning algorithms to detect certain morphological structures, e.g. cell nuclei, that are associated with certain stains, improving the result of the normalization process~(\cite{khan2014nonlinear,bejnordi2016stain}). Deep generative models, i.e. variational autoencoders and generative adversarial networks~(\cite{kingma2013auto, goodfellow2014generative}), have been used to generate new image samples that match the template data manifold~(\cite{cho2017neural,zanjani2018stain}). Moreover, color normalization has been formulated as a style transfer task where the style is defined as the color distribution produced by a particular lab~(\cite{bug2017context}). However, despite their success and widespread adoption as a preprocessing tool in a variety of computational pathology applications~(\cite{clarke2017colour,albarqouni2016aggnet,janowczyk2017stain,bandi2018detection}), they are not always effective and can produce images with color distributions that deviate from the desired color template. In this study, we propose a novel unsupervised approach that leverages the power of deep learning to solve the problem of stain normalization. We reformulate the problem of stain normalization as an image-to-image translation task and train a neural network to solve it. We do so by feeding the network with heavily augmented H\&E images and training the model to reconstruct the original image without augmentation. By learning to remove this color variation, the network effectively learns to perform \textit{stain color normalization} in unseen images whose color distribution deviates from that of the training set.

\subsection{Multicenter evaluation}

Despite the wide adoption of \textit{stain color augmentation} and \textit{stain color normalization} in the field of computational pathology, the effects on performance of these techniques have not been systematically evaluated. Existing literature focuses on particular applications, and does not quantify the relationship between these techniques and CNN performance~(\cite{komura2018machine,wang2015exploring,zhu2014scalable,veta2018predicting}). In this study, we aim to overcome this limitation by comparing these techniques across four representative applications including multicenter data. We selected four patch-based classification tasks where a CNN was trained with data from a single center only, and evaluated in unseen data from multiple external pathology laboratories. We chose four relevant applications from the literature: (1) detecting the presence of mitotic figures in breast tissue~(\cite{tellez2018whole}); (2) detecting the presence of tumor metastases in breast lymph node tissue~(\cite{bandi2018detection}); (3) detecting the presence of epithelial cells in prostate tissue~(\cite{bulten2019epithelium}); and (4) distinguishing among 9 tissue classes in colorectal cancer (CRC) tissue~(\cite{ciompi2017importance}). All test datasets presented a substantial and challenging stain color deviation from the training set, as can be seen in Fig.~\ref{fig:patches}. We trained a series of CNN classifiers following an identical training protocol while varying the \textit{stain color normalization} and \textit{stain color augmentation} techniques used during training. This thorough evaluation allowed us to establish a ranking among the methods and measure relative performance improvements among them.  

\subsection{Contributions}

Our contributions can be summarized as follows:

\begin{itemize}
  \item We systematically evaluated several well-known \textit{stain color augmentation} and \textit{stain color normalization} algorithms in order to quantify their effects on CNN classification performance.
  \item We conducted the previous evaluation using data from a total of 9 different centers spanning 4 relevant classification tasks: mitosis detection, tumor metastasis detection in lymph nodes, prostate epithelium detection, and multiclass colorectal cancer tissue classification.
  \item We formulated the problem of \textit{stain color normalization} as an unsupervised image-to-image translation task and trained a neural network to solve it. 
 \end{itemize}

The paper is organized as follows. Sec. \ref{sec:materials} and Sec. \ref{sec:methods} describe the materials and methods thoroughly. Experimental results are explained in Sec. \ref{sec:experiments}, followed by Sec. \ref{sec:discussion} and Sec. \ref{sec:conclusion} where the discussion and final conclusion are stated.

\section{Materials}
\label{sec:materials}

We collected data from a variety of pathology laboratories for four different applications. In all cases, we used images from the Radboud University Medical Centre (Radboudumc or \textit{rumc}) exclusively to train the models for each of the four classification tasks. Images from the remaining centers were used for testing purposes only. We considered RGB patches of 128x128 pixels extracted from annotated regions. Examples of these patches are shown in Fig.~\ref{fig:patches}. The following sections describe each of the four classification tasks.

\subsection{Mitotic figure detection}

In this binary classification task, the goal was to accurately classify as positive samples those patches containing a mitotic figure in their center, i.e. a cell undergoing division. In order to train the classifier, we used 14 H\&E WSIs from triple negative breast cancer patients, scanned at \SI{0.25}{\um/pixel} resolution, with annotations of mitotic figures obtained as described in~(\cite{tellez2018whole}). We split the slides into training (6), validation (4) and test (4), and extracted a total of 1M patches. We refer to this set as \textit{mitosis-rumc}.

For the external dataset, we used publicly available data from the TUPAC Challenge~(\cite{veta2018predicting}), i.e. 50 cases of invasive breast cancer with manual annotations of mitotic figures scanned at \SI{0.25}{\um/pixel} resolution. We extracted a total of 300K patches, and refer to this dataset as \textit{mitosis-tupac}.

\subsection{Tumor metastasis detection}

The aim of this binary classification task was to identify patches containing metastatic tumor cells. We used publicly available WSIs from the Camelyon17 Challenge~(\cite{bandi2018detection}). This cohort consisted of 50 exhaustively annotated H\&E slides of breast lymph node resections from breast cancer patients from 5 different centers (10 slides per center), including Radboudumc. They were scanned at \SI{0.25}{\um/pixel} resolution and the tumor metastases were manually delineated by experts. 

We used the 10 WSIs from the Radboudumc to train the classifier, split into training (4), validation (3) and test (3), and extracted a total of 300K patches. We refer to this dataset as \textit{lymph-rumc}. We used the remaining 40 WSIs as external test data, extracting a total of 1.2M patches, and assembling 4 different test sets (one for each center). We named them according to their center's name acronym: \textit{lymph-umcu}, \textit{lymph-cwh}, \textit{lymph-rh} and \textit{lymph-lpe}.

\begin{figure*}[!h]
\centering
\includegraphics[width=0.9\textwidth]{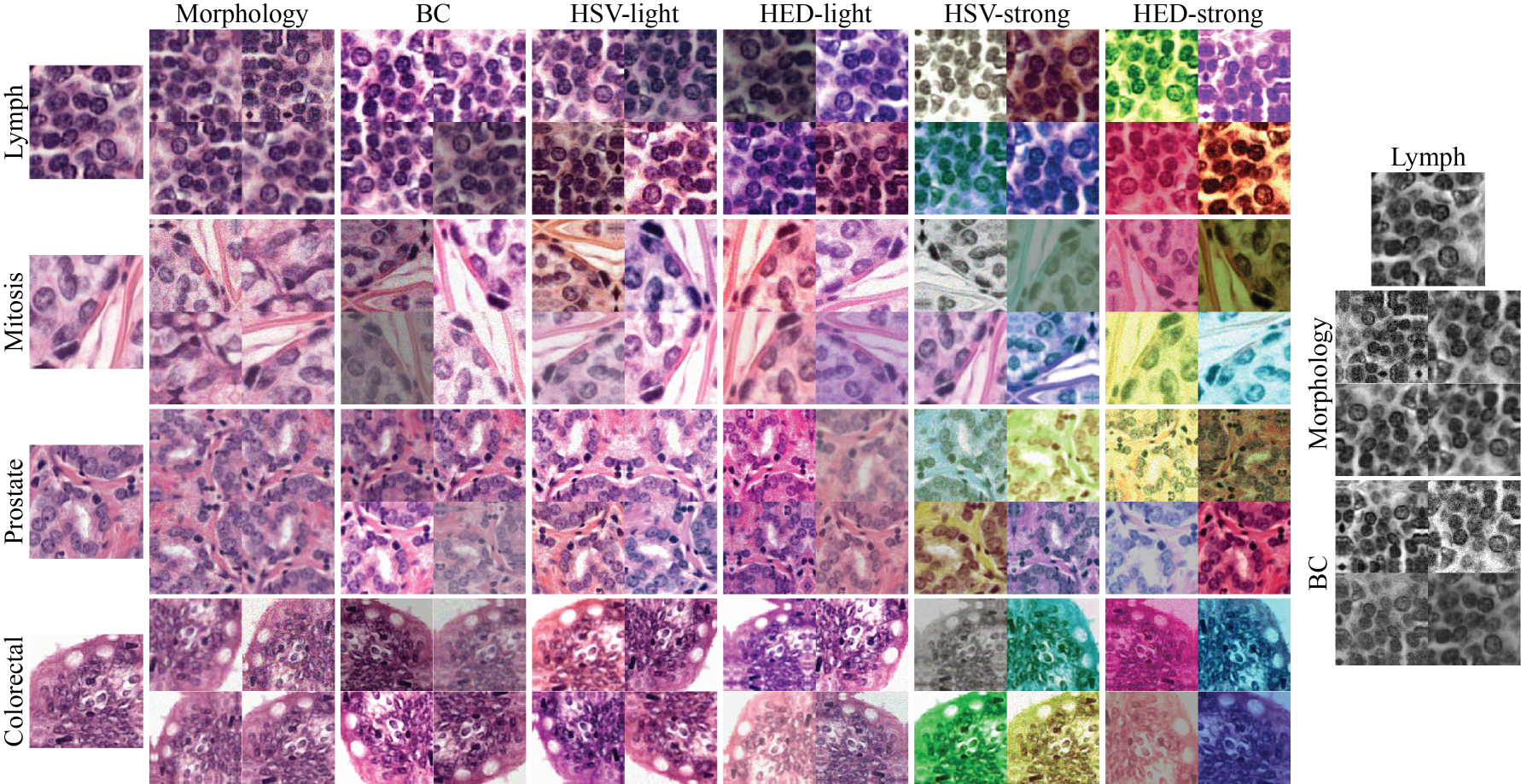}
\caption{\label{fig:augmentation} Summary of the data augmentation techniques and datasets used in this study, organized in columns and rows respectively. Patches on the leftmost column depict the original input images and the rest of patches are augmented versions of them. Augmentations performed in the \textit{grayscale} color space are depicted on the right for one sample dataset only. \textit{Basic} augmentation is included in all cases.}
\end{figure*}

\subsection{Prostate epithelium detection}

The goal of this binary classification task was to identify patches containing epithelial cells in prostate tissue. We trained the classifier with 25 H\&E WSIs of prostate resections from the Radboudumc scanned at \SI{0.5}{\um/pixel} resolution, with annotations of epithelial tissue as described in~(\cite{bulten2019epithelium}). We split this cohort into training (13), validation (6) and test (6), and extracted a total of 250K patches. We refer to it as \textit{prostate-rumc}.

We used two test datasets for this task. First, we selected 10 H\&E slides of prostate resections from the Radboudumc with different staining and scanning conditions, resulting in substantially different stain appearance (see \textit{prostate-rumc2} in Fig.~\ref{fig:patches}). This test set was manually annotated as described in~(\cite{bulten2019epithelium}) and named \textit{prostate-rumc2}. We extracted 75K patches from these WSIs. Second, we used publicly available images from 20 H\&E slides of prostatectomy specimens with manual annotations of epithelial tissue obtained as described in~(\cite{bulten2019epithelium,gertych2015machine}). We extracted 65K patches from them and named the test set \textit{prostate-cedar}.

\subsection{Colorectal cancer tissue type classification}

In this multiclass classification task, the goal was to distinguish among 9 different colorectal cancer (CRC) tissue classes, namely: 1)~tumor, 2)~stroma, 3)~muscle, 4)~lymphocytes, 5)~healthy glands, 6)~fat, 7)~blood cells, 8)~necrosis and debris, and 9)~mucus. We used 54 H\&E WSIs of colorectal carcinoma tissue from the Radboudumc scanned at \SI{0.5}{\um/pixel} resolution to train the classifier, with manual annotations of the 9 different tissue classes. We split this cohort into training (24), validation (15) and test (15), extracted a total of 450K patches, and named it \textit{crc-rumc}.

We used two external datasets for this task. First, a set of 74 H\&E WSIs from rectal carcinoma patients with annotations of the same 9 classes, as described in~(\cite{ciompi2017importance}). We extracted 35K patches and refer to this dataset as \textit{crc-labpon}. Second, we used a publicly available set of H\&E image patches from colorectal carcinoma patients~(\cite{kather2016multi}). Annotations for 6 tissue types were available: 1)~tumor, 2)~stroma, 3)~lymph, 4)~healthy glands, 5)~fat, and 6)~blood cells, debris and mucus. We extracted 4K patches in total, and refer to this dataset as \textit{crc-heidelberg}.

\subsection{Multi-organ dataset}

For the purpose of training a network to solve the problem of \textit{stain color normalization}, we created an auxiliary dataset by aggregating patches from \textit{mitosis-rumc}, \textit{lymph-rumc}, \textit{prostate-rumc} and \textit{crc-rumc} in a randomized and balanced manner. We discarded all labels since they were not needed for this purpose. We preserved a total of 500K patches for this set and called it the \textit{multi-organ} dataset.

\section{Methods}
\label{sec:methods}

In this study, we evaluated the effect in classification performance of several methods for \textit{stain color augmentation} and \textit{stain color normalization}. This section describes these methods.

\begin{figure*}[t]
\centering
\includegraphics[width=0.8\textwidth]{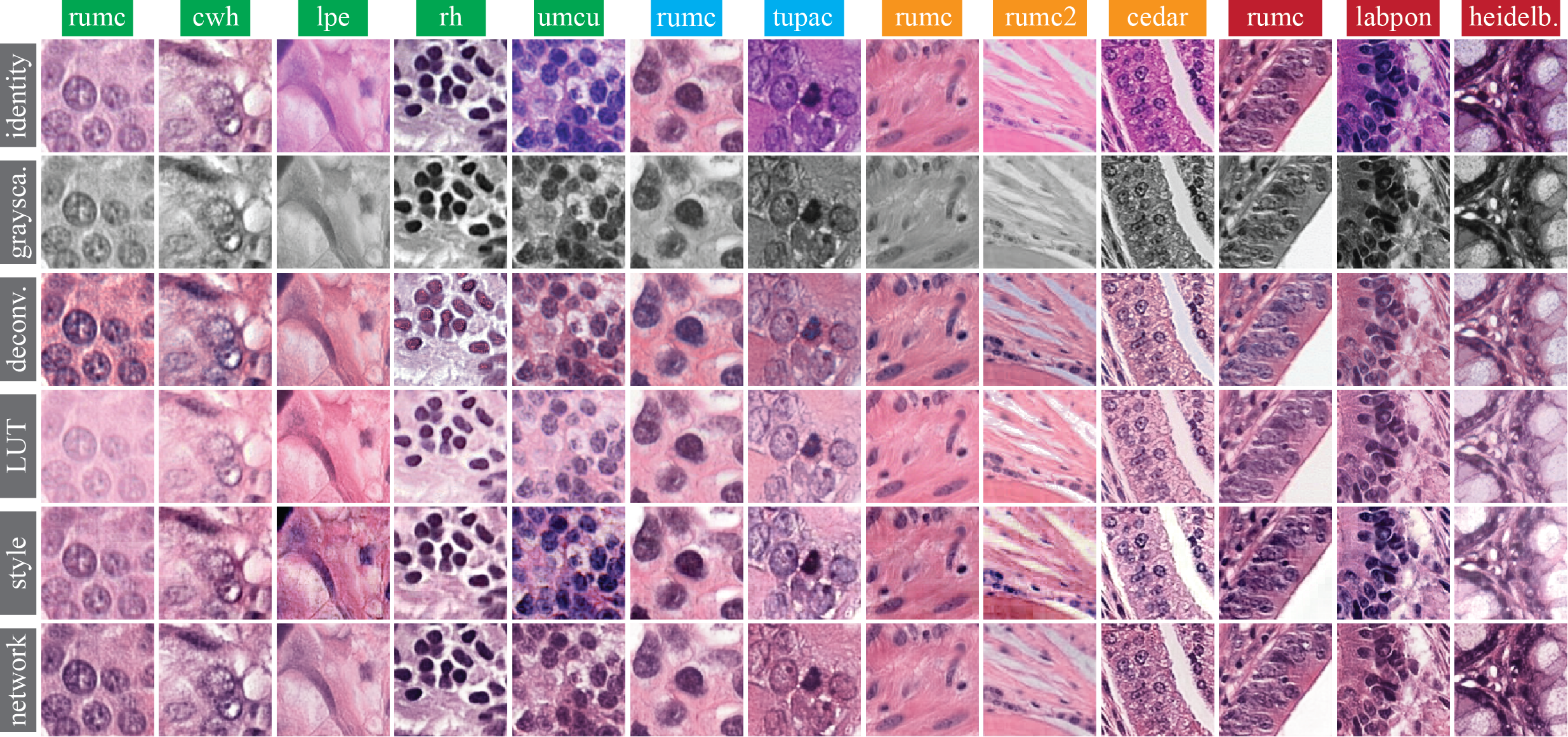}
\caption{\label{fig:patches_std} Visual comparison of the stain color normalization techniques used in this study. Rows correspond to the different tested techniques and columns to datasets, with green for \textit{lymph}, blue for \textit{mitosis}, yellow for \textit{prostate} and red for \textit{colorectal}.}
\end{figure*}

\subsection{Stain color augmentation}

We assume a homogeneous stain color distribution $\phi_{\text{train}}$ for the training images and a more varied color distribution $\phi_{\text{test}}$ for the test images. Note that it is challenging for a classification model trained solely with $\phi_{\text{train}}$ to generalize well to $\phi_{\text{test}}$ due to potential stain differences among sets. To solve this problem, stain color augmentation defines a preprocessing function $f$ that transforms images of the training set to present an alternative and more diverse color distribution $\phi_{\text{augment}}$:

\begin{equation}
\label{eq:augment}
\phi_{\text{train}} \xrightarrow{f} \phi_{\text{augment}}
\end{equation}

on the condition that:

\begin{equation}
\label{eq:augment_condition}
(\phi_{\text{augment}} \supseteq \phi_{\text{train}}) \land (\phi_{\text{augment}} \supseteq \phi_{\text{test}})
\end{equation}

In practice, heavy data augmentation is used to satisfy Eq.~\ref{eq:augment_condition}. In order to simplify our experimental setup, we grouped several data augmentation techniques into the following categories attending to the nature of the image transformations. Examples of the resulting augmented images are shown in Fig.~\ref{fig:augmentation}.

\textbf{Basic}. This group included 90 degree rotations, and vertical and horizontal mirroring. 

\textbf{Morphology}. We extended \textit{basic} with several transformations that simulate morphological perturbations, i.e. alterations in shape, texture or size of the imaged tissue structures, including scanning artifacts. We included \textit{basic} augmentation, scaling, elastic deformation~(\cite{simard2003best}), additive Gaussian noise (perturbing the signal-to-noise ratio), and Gaussian blurring (simulating out-of-focus artifacts).

\textbf{Brightness \& contrast (BC)}. We extended \textit{morphology} with random brightness and contrast image perturbations~(\cite{haeberli1994image}).

\textbf{Hue-Saturation-Value (HSV)}. We extended the \textit{BC} augmentation by randomly shifting the hue and saturation channels in the HSV color space~(\cite{van2014scikit}). This transformation produced substantially different color distributions when applied to the training images. We tested two configurations depending on the observed color variation strength, called \textit{HSV-light} and \textit{HSV-strong}.

\textbf{Hematoxylin-Eosin-DAB (HED)}. We extended the \textit{BC} augmentation with a color variation routine specifically designed for H\&E images~(\cite{tellez2018whole}). This method followed three steps. First, it disentangled the hematoxylin and eosin color channels by means of color deconvolution using a fixed matrix. Second, it perturbed the hematoxylin and eosin stains independently. Third, it transformed the resulting stains into regular RGB color space. We tested two configurations depending on the observed color variation strength, called \textit{HED-light} and \textit{HED-strong}.

During training, we selected the value of the augmentation hyper-parameters randomly within certain ranges to achieve stain variation. We tuned all ranges manually via visual examination. In particular, we used a scaling factor between~$[0.8, 1.2]$, elastic deformation parameters $\alpha\in[80, 120]$ and $\sigma\in[9.0, 11.0]$, additive Gaussian noise with~$\sigma\in[0,0.1]$, Gaussian blurring with~$\sigma\in[0,0.1]$, brightness intensity ratio between~$[0.65, 1.35]$, and contrast intensity ratio between~$[0.5, 1.5]$. For \textit{HSV-light} and \textit{HSV-strong}, we used hue and saturation intensity ratios between~$[-0.1, 0.1]$ and~$[-1, 1]$, respectively. For \textit{HED-light} and \textit{HED-strong}, we used intensity ratios between~$[-0.05, 0.05]$ and~$[-0.2, 0.2]$, respectively, for all HED channels.

\subsection{Stain color normalization}

\begin{figure}[!t]
\centering
\includegraphics[width=0.47\textwidth]{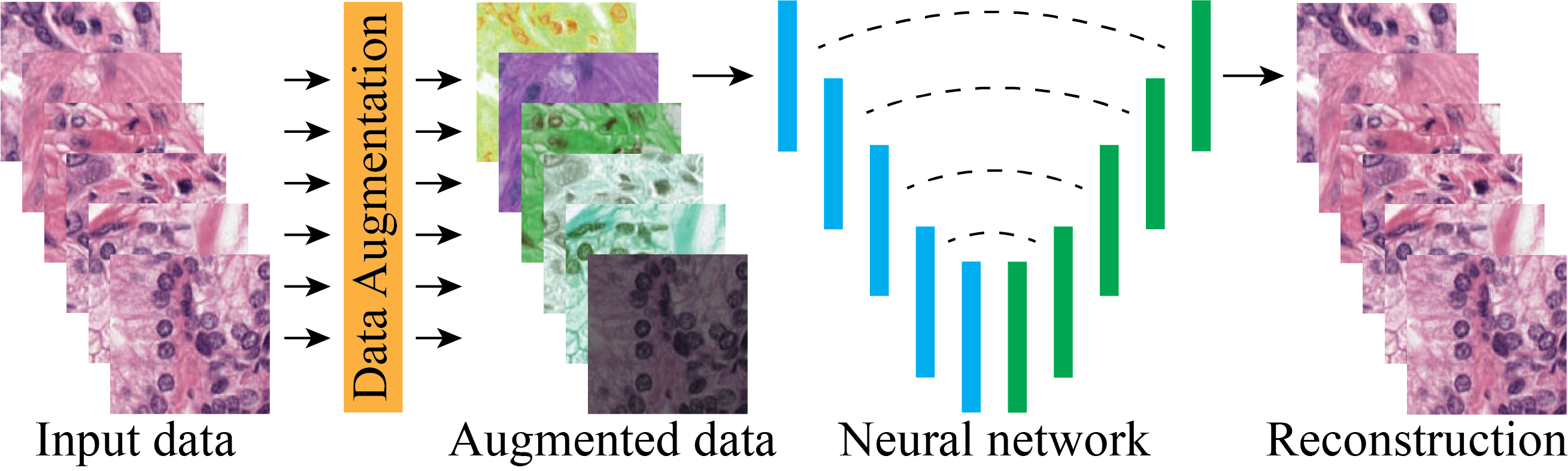}
\caption{\label{fig:network-std} Network-based stain color normalization. From left to right: patches from the training set are transformed with heavy color augmentation and fed to a neural network. This network is trained to reconstruct the original appearance of the input images by removing color augmentation, effectively learning how to perform stain color normalization.}
\end{figure}

Stain color normalization reduces color variation by transforming the color distribution of training and test images, i.e. $\phi_{\text{train}}$ and $\phi_{\text{test}}$, to that of a template $\phi_{\text{normal}}$. It performs such transformation using a normalization function $g$ that maps any given color distribution to the template one:

\begin{equation}
\label{eq:std}
(\phi_{\text{train}} \xrightarrow{g} \phi_{\text{normal}}) \land (\phi_{\text{test}} \xrightarrow{g} \phi_{\text{normal}})
\end{equation}

By matching $\phi_{\text{train}}$ and $\phi_{\text{test}}$, the problem of stain variance vanishes and the model no longer requires to generalize to unseen stains in order to perform well. We evaluated several methods that implement $g$ (see Fig.~\ref{fig:patches_std}), and propose a novel technique based on neural networks.

\begin{figure*}[h!]
\centering
\includegraphics[width=0.9\textwidth]{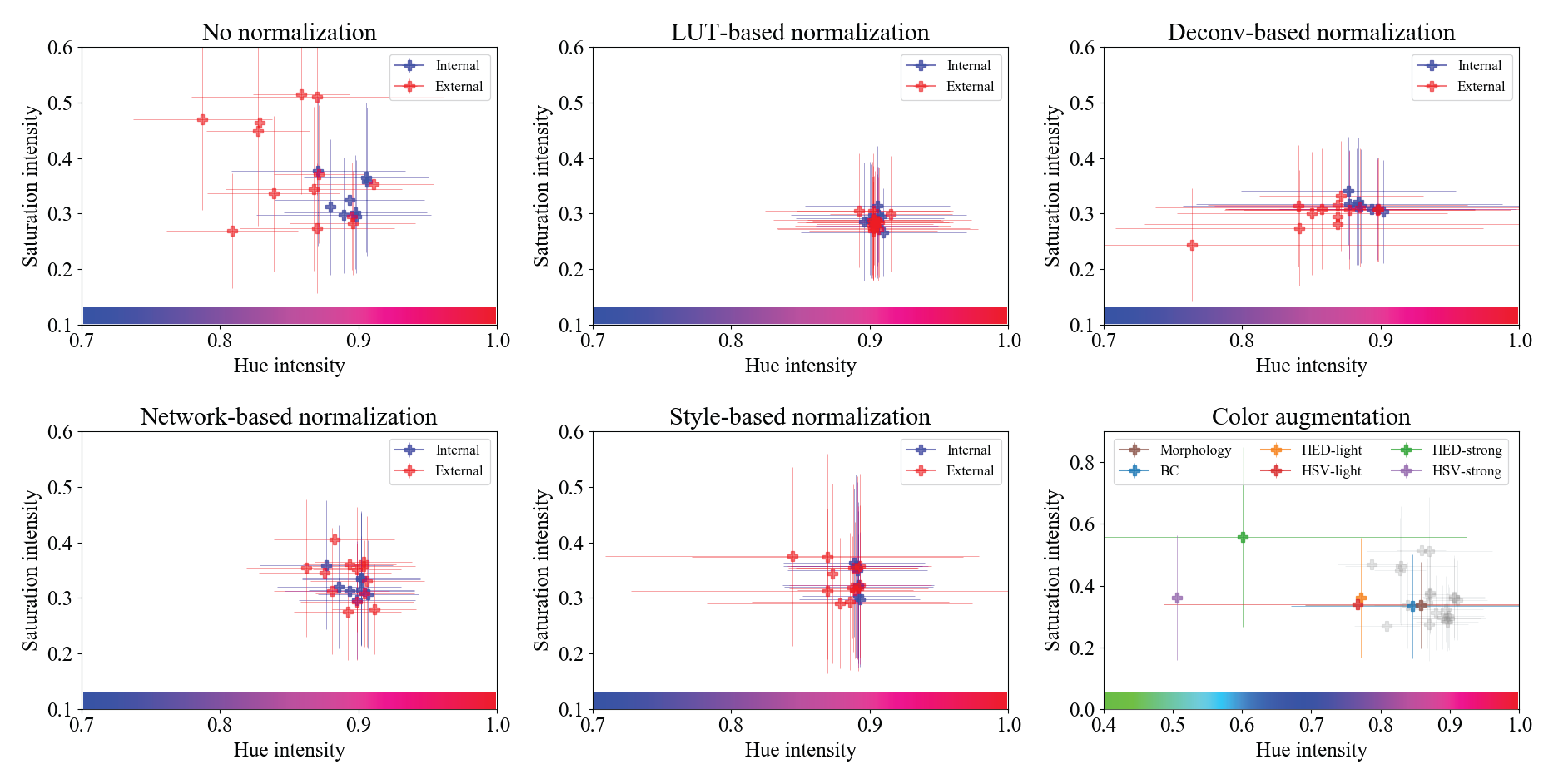}
\caption{\label{fig:color_space} Constellations of internal and external datasets analyzed in this work. Each data point represents the mean and standard deviation pixel intensity of all image patches in a particular dataset in the HSV color space (hue and saturation in the x and y axis, respectively). Note how normalization methods tend to cluster the color distribution of the datasets, whereas color augmentation does the opposite. Color augmentation plot (bottom-right): patches from internal images are transformed with different color augmentation methods (grey points representing the original internal and external datasets are shown as reference).}
\end{figure*}

\textbf{Identity}. We performed no transformation on the input patches, serving as a baseline method for the rest of techniques. 

\textbf{Grayscale}. In this case, $g$ transformed images from RGB to grayscale space, removing most of the color information present in the patches. We hypothesized that this color information is redundant since most of the signal in H\&E images is present in morphological and structural patterns, e.g. the presence of a certain type of cell. 

\textbf{Deconv-based}. We followed the color deconvolution approach proposed by~(\cite{macenko2009method}). This method assumes that the hematoxylin and eosin stains are linearly separable in the optical density (OD) color space, as opposed to RGB space. This method finds the two largest singular value directions using singular value decomposition, and projects the OD pixel values onto this plane. This procedure allows to identify the underlying hematoxylin and eosin stain vectors, and use them to perform color deconvolution on a given image to decompose the RGB image into its normalized hematoxylin and eosin components. 

\textbf{LUT-based}. We implemented an approach that uses tissue morphology to perform stain color normalization~(\cite{bejnordi2016stain}). This popular method has been used by numerous researchers in recent public challenges~(\cite{bandi2018detection, bejnordi2017diagnostic}). It detects cell nuclei in order to precisely characterize the H\&E chromatic distribution and density histogram for a given WSI. First, it does so for a given template WSI, e.g. an image from the training set, and a target WSI. Second, the color distributions of the template and target WSIs are matched, and the color correspondence is stored in a look-up table (LUT). Finally, this LUT is used to normalize the color of the target WSI.

\textbf{Style-based}. \cite{bug2017context} proposed to use a neural network to perform stain color normalization based on the idea of style transfer. They transform the color distribution of RGB images by using feature-aware normalization, a mechanism that shifts and scales intermediate feature maps based on features extracted from the input image. This feature extractor is an ImageNet~(\cite{deng2009imagenet}) pre-trained network, while the rest of the model is trained to reconstruct PCA-augmented histopathology images. We used the authors' implementation of the method and retrained the model using images from the \textit{multi-organ} dataset.

\textbf{Network-based}. We developed a novel approach to perform stain color normalization based on unsupervised learning and neural networks (see Fig.~\ref{fig:network-std}). We parameterized the normalization function $g$ with a neural network $G$ and trained it end-to-end to remove the effects of data augmentation. Even though it is not possible to invert the many-to-many augmentation function $f$, we can learn a partial many-to-one function that maps any arbitrary color distribution $\phi_{\text{augment}}$ to a template distribution $\phi_{\text{normal}}$: 

\begin{equation}
\label{eq:network-std}
\phi_{\text{augment}} \xrightarrow{G} \phi_{\text{normal}}
\end{equation}

Since $G$ can normalize $\phi_{\text{augment}}$ (Eq.~\ref{eq:network-std}), and  $\phi_{\text{augment}}$ is a superset of $\phi_{\text{train}}$ and $\phi_{\text{test}}$ (Eq.~\ref{eq:augment_condition}), we conclude that $G$ can effectively normalize $\phi_{\text{train}}$ and $\phi_{\text{test}}$ (Eq.~\ref{eq:augment_condition}).

We trained $G$ to perform image-to-image translation using the \textit{multi-organ} dataset. During training, images were heavily augmented and fed to the network. The model was tasked with reconstructing the images with their original appearance, before augmentation. We used a special configuration of the \textit{HSV} augmentation where we kept the color transformation only, i.e. did not include \textit{basic}, \textit{morphology} and \textit{BC}. We used the maximum intensity for the transformation hyper-parameters, i.e. hue, saturation and value channel ratios between ~$[-1, 1]$. The strength of this augmentation resulted in images with drastically different color distributions, sometimes compressing all color information into grayscale. In order to invert this complex augmentation, we hypothesized that the network learned to associate certain tissue structures with their usual color appearance.

\begin{sidewaystable*}[ph!]

\scriptsize
\caption{\label{tab:ranking} Experimental results ranking \textit{stain color augmentation} and \textit{stain color normalization} methods. Values correspond to AUC scores, except for the last column, averaged across 5 repetitions with standard deviation shown between parenthesis. Each column represents a different external test dataset, with the last column \textit{Ranking} indicating the position of each method within the global benchmark, computed as described in Sec.~\ref{sec:experiments_evaluation}. Normalization methods: \textit{Network} is our proposal; \textit{Style} is from~\cite{bug2017context}; \textit{LUT} is from~\cite{bejnordi2016stain}; and \textit{Deconvolution} is from~\cite{macenko2009method}.}
\centering
\begin{tabular}{|ll|llll|l|ll|ll|l|}
\hline
\multicolumn{1}{|c}{\textbf{Normalization}} & \multicolumn{1}{c|}{\textbf{Augmentation}} & \multicolumn{1}{c}{\textbf{lymph-cwh}} & \multicolumn{1}{c}{\textbf{lymph-lpe}} & \multicolumn{1}{c}{\textbf{lymph-rh}} & \multicolumn{1}{c|}{\textbf{lymph-umcu}} & \multicolumn{1}{c|}{\textbf{mitosis-tupac}} & \multicolumn{1}{c}{\textbf{prostate-rumc2}} & \multicolumn{1}{c|}{\textbf{prostate-cedar}} & \multicolumn{1}{c}{\textbf{crc-labpon}} & \multicolumn{1}{c|}{\textbf{crc-heidelberg}} & \multicolumn{1}{c|}{\textbf{Ranking}} \\ \hline
Identity                                    & HED-light                                  & 0.952(0.004)                           & 0.976(0.001)                           & 0.946(0.009)                          & 0.968(0.004)                             & 0.996(0.001)                                & 0.957(0.001)                                & 0.879(0.011)                                 & 0.973(0.002)                            & 0.895(0.002)                                 & 1.2(0.4)                              \\
Style                                       & HED-light                                  & 0.961(0.002)                           & 0.953(0.004)                           & 0.952(0.001)                          & 0.972(0.004)                             & 0.991(0.003)                                & 0.925(0.003)                                & 0.879(0.006)                                 & 0.975(0.001)                            & 0.917(0.003)                                 & 2.8(0.7)                              \\
Network                                     & HSV-light                                  & 0.946(0.006)                           & 0.962(0.001)                           & 0.941(0.002)                          & 0.965(0.004)                             & 0.992(0.001)                                & 0.957(0.000)                                & 0.872(0.013)                                 & 0.980(0.001)                            & 0.900(0.003)                                 & 3.9(1.9)                              \\
Network                                     & HED-light                                  & 0.949(0.005)                           & 0.968(0.001)                           & 0.942(0.002)                          & 0.963(0.004)                             & 0.989(0.003)                                & 0.958(0.001)                                & 0.862(0.011)                                 & 0.980(0.001)                            & 0.906(0.003)                                 & 4.1(1.6)                              \\
Identity                                    & HSV-strong                                 & 0.955(0.003)                           & 0.965(0.004)                           & 0.929(0.002)                          & 0.973(0.003)                             & 0.988(0.003)                                & 0.945(0.009)                                & 0.886(0.005)                                 & 0.977(0.001)                            & 0.902(0.003)                                 & 4.7(1.7)                              \\
Network                                     & HSV-strong                                 & 0.953(0.002)                           & 0.964(0.003)                           & 0.946(0.002)                          & 0.964(0.005)                             & 0.991(0.003)                                & 0.951(0.002)                                & 0.852(0.006)                                 & 0.975(0.002)                            & 0.894(0.005)                                 & 6.6(0.9)                              \\
Network                                     & HED-strong                                 & 0.956(0.003)                           & 0.959(0.002)                           & 0.940(0.003)                          & 0.965(0.004)                             & 0.985(0.005)                                & 0.943(0.003)                                & 0.861(0.009)                                 & 0.974(0.002)                            & 0.916(0.003)                                 & 7.9(1.9)                              \\
Identity                                    & HED-strong                                 & 0.950(0.005)                           & 0.959(0.005)                           & 0.936(0.007)                          & 0.957(0.007)                             & 0.992(0.002)                                & 0.945(0.003)                                & 0.872(0.005)                                 & 0.967(0.003)                            & 0.920(0.005)                                 & 8.1(2.8)                              \\
Style                                       & HSV-strong                                 & 0.953(0.004)                           & 0.956(0.004)                           & 0.940(0.003)                          & 0.959(0.007)                             & 0.986(0.004)                                & 0.932(0.005)                                & 0.878(0.003)                                 & 0.976(0.001)                            & 0.917(0.004)                                 & 9.0(2.6)                              \\
Style                                       & HSV-light                                  & 0.940(0.011)                           & 0.960(0.004)                           & 0.944(0.007)                          & 0.926(0.012)                             & 0.992(0.001)                                & 0.958(0.002)                                & 0.852(0.008)                                 & 0.974(0.001)                            & 0.921(0.003)                                 & 9.4(3.6)                              \\
Style                                       & HED-strong                                 & 0.955(0.002)                           & 0.949(0.004)                           & 0.936(0.003)                          & 0.954(0.005)                             & 0.982(0.005)                                & 0.942(0.002)                                & 0.884(0.004)                                 & 0.975(0.000)                            & 0.925(0.003)                                 & 9.9(1.2)                              \\
Grayscale                                   & BC                                         & 0.956(0.003)                           & 0.962(0.003)                           & 0.935(0.005)                          & 0.961(0.002)                             & 0.989(0.002)                                & 0.939(0.004)                                & 0.851(0.002)                                 & 0.972(0.000)                            & 0.884(0.003)                                 & 12.2(1.2)                             \\
Deconvolution                               & HSV-strong                                 & 0.955(0.003)                           & 0.936(0.008)                           & 0.941(0.004)                          & 0.943(0.009)                             & 0.991(0.001)                                & 0.865(0.010)                                & 0.867(0.004)                                 & 0.961(0.002)                            & 0.928(0.001)                                 & 13.9(1.9)                             \\
LUT                                         & HED-strong                                 & 0.934(0.006)                           & 0.941(0.006)                           & 0.925(0.006)                          & 0.963(0.005)                             & 0.989(0.002)                                & 0.945(0.002)                                & 0.871(0.005)                                 & 0.956(0.002)                            & 0.945(0.001)                                 & 14.0(2.2)                             \\
Deconvolution                               & HED-strong                                 & 0.942(0.003)                           & 0.962(0.003)                           & 0.897(0.006)                          & 0.967(0.003)                             & 0.993(0.002)                                & 0.827(0.018)                                & 0.853(0.006)                                 & 0.969(0.001)                            & 0.927(0.002)                                 & 14.4(1.2)                             \\
LUT                                         & HSV-strong                                 & 0.923(0.009)                           & 0.939(0.003)                           & 0.928(0.005)                          & 0.947(0.008)                             & 0.987(0.002)                                & 0.949(0.003)                                & 0.862(0.007)                                 & 0.962(0.002)                            & 0.940(0.002)                                 & 17.0(2.0)                             \\
Network                                     & BC                                         & 0.944(0.003)                           & 0.950(0.003)                           & 0.903(0.003)                          & 0.934(0.006)                             & 0.983(0.005)                                & 0.953(0.003)                                & 0.869(0.009)                                 & 0.981(0.001)                            & 0.881(0.005)                                 & 17.4(1.5)                             \\
Identity                                    & HSV-light                                  & 0.888(0.013)                           & 0.951(0.009)                           & 0.942(0.004)                          & 0.930(0.023)                             & 0.962(0.015)                                & 0.949(0.001)                                & 0.905(0.005)                                 & 0.976(0.000)                            & 0.894(0.003)                                 & 17.4(2.9)                             \\
LUT                                         & HED-light                                  & 0.914(0.011)                           & 0.926(0.011)                           & 0.923(0.006)                          & 0.932(0.019)                             & 0.993(0.001)                                & 0.948(0.003)                                & 0.852(0.021)                                 & 0.966(0.003)                            & 0.940(0.002)                                 & 17.9(2.2)                             \\
LUT                                         & HSV-light                                  & 0.894(0.006)                           & 0.936(0.006)                           & 0.921(0.003)                          & 0.942(0.007)                             & 0.987(0.002)                                & 0.951(0.002)                                & 0.860(0.010)                                 & 0.971(0.002)                            & 0.945(0.002)                                 & 19.2(1.2)                             \\
LUT                                         & BC                                         & 0.925(0.025)                           & 0.948(0.027)                           & 0.853(0.016)                          & 0.790(0.061)                             & 0.985(0.004)                                & 0.951(0.004)                                & 0.848(0.018)                                 & 0.973(0.003)                            & 0.924(0.005)                                 & 21.3(3.3)                             \\
Style                                       & BC                                         & 0.949(0.005)                           & 0.858(0.031)                           & 0.938(0.001)                          & 0.411(0.065)                             & 0.987(0.004)                                & 0.949(0.006)                                & 0.764(0.047)                                 & 0.946(0.002)                            & 0.903(0.005)                                 & 23.3(2.2)                             \\
Deconvolution                               & HSV-light                                  & 0.942(0.004)                           & 0.930(0.009)                           & 0.913(0.023)                          & 0.961(0.002)                             & 0.982(0.005)                                & 0.850(0.019)                                & 0.840(0.009)                                 & 0.958(0.006)                            & 0.917(0.002)                                 & 23.5(1.0)                             \\
Network                                     & Basic                                   & 0.944(0.003)                           & 0.954(0.007)                           & 0.887(0.010)                          & 0.959(0.004)                             & 0.969(0.005)                                & 0.905(0.006)                                & 0.815(0.019)                                 & 0.977(0.002)                            & 0.855(0.006)                                 & 23.6(1.4)                             \\
Network                                     & Morphology                                 & 0.939(0.010)                           & 0.949(0.006)                           & 0.890(0.012)                          & 0.950(0.009)                             & 0.980(0.006)                                & 0.913(0.011)                                & 0.823(0.022)                                 & 0.977(0.001)                            & 0.868(0.002)                                 & 23.9(1.1)                             \\
Deconvolution                               & HED-light                                  & 0.930(0.005)                           & 0.912(0.015)                           & 0.916(0.005)                          & 0.948(0.006)                             & 0.982(0.002)                                & 0.816(0.011)                                & 0.834(0.004)                                 & 0.970(0.003)                            & 0.927(0.005)                                 & 25.4(2.3)                             \\
Deconvolution                               & Morphology                                 & 0.951(0.003)                           & 0.938(0.006)                           & 0.849(0.021)                          & 0.951(0.008)                             & 0.993(0.002)                                & 0.754(0.027)                                & 0.749(0.037)                                 & 0.903(0.008)                            & 0.865(0.015)                                 & 27.7(0.6)                             \\
Grayscale                                   & Morphology                                 & 0.943(0.010)                           & 0.820(0.021)                           & 0.922(0.005)                          & 0.941(0.011)                             & 0.991(0.006)                                & 0.910(0.009)                                & 0.816(0.005)                                 & 0.929(0.006)                            & 0.813(0.009)                                 & 27.7(1.2)                             \\
Style                                       & Morphology                                 & 0.935(0.011)                           & 0.725(0.082)                           & 0.934(0.002)                          & 0.361(0.113)                             & 0.992(0.004)                                & 0.918(0.006)                                & 0.754(0.006)                                 & 0.873(0.010)                            & 0.890(0.006)                                 & 28.6(2.4)                             \\
Grayscale                                   & Basic                                   & 0.940(0.007)                           & 0.692(0.064)                           & 0.926(0.010)                          & 0.938(0.019)                             & 0.992(0.001)                                & 0.882(0.008)                                & 0.661(0.039)                                 & 0.934(0.002)                            & 0.798(0.006)                                 & 30.0(0.6)                             \\
Deconvolution                               & BC                                         & 0.942(0.004)                           & 0.896(0.008)                           & 0.682(0.044)                          & 0.949(0.005)                             & 0.989(0.006)                                & 0.794(0.021)                                & 0.792(0.028)                                 & 0.930(0.007)                            & 0.872(0.004)                                 & 30.4(1.2)                             \\
LUT                                         & Morphology                                 & 0.898(0.007)                           & 0.920(0.007)                           & 0.801(0.021)                          & 0.874(0.025)                             & 0.969(0.008)                                & 0.895(0.013)                                & 0.803(0.007)                                 & 0.939(0.006)                            & 0.906(0.006)                                 & 32.6(1.4)                             \\
Deconvolution                               & Basic                                   & 0.919(0.015)                           & 0.896(0.038)                           & 0.810(0.081)                          & 0.902(0.026)                             & 0.993(0.001)                                & 0.753(0.006)                                & 0.791(0.009)                                 & 0.903(0.003)                            & 0.836(0.008)                                 & 32.8(0.7)                             \\
Style                                       & Basic                                   & 0.918(0.002)                           & 0.334(0.133)                           & 0.926(0.004)                          & 0.124(0.041)                             & 0.991(0.003)                                & 0.865(0.025)                                & 0.723(0.024)                                 & 0.863(0.020)                            & 0.857(0.010)                                 & 33.6(1.0)                             \\
LUT                                         & Basic                                   & 0.908(0.010)                           & 0.894(0.030)                           & 0.809(0.022)                          & 0.772(0.072)                             & 0.951(0.009)                                & 0.906(0.011)                                & 0.741(0.018)                                 & 0.930(0.014)                            & 0.890(0.013)                                 & 34.6(0.8)                             \\
Identity                                    & BC                                         & 0.899(0.006)                           & 0.634(0.100)                           & 0.741(0.016)                          & 0.177(0.047)                             & 0.906(0.034)                                & 0.936(0.006)                                & 0.704(0.060)                                 & 0.684(0.009)                            & 0.761(0.012)                                 & 36.2(0.4)                             \\
Identity                                    & Morphology                                 & 0.811(0.026)                           & 0.671(0.099)                           & 0.673(0.027)                          & 0.214(0.174)                             & 0.986(0.006)                                & 0.374(0.191)                                & 0.602(0.023)                                 & 0.569(0.028)                            & 0.720(0.009)                                 & 37.2(0.7)                             \\
Identity                                    & Basic                                   & 0.811(0.009)                           & 0.563(0.309)                           & 0.790(0.047)                          & 0.406(0.375)                             & 0.965(0.009)                                & 0.631(0.178)                                & 0.624(0.053)                                 & 0.556(0.057)                            & 0.701(0.028)                                 & 37.6(0.5)                             \\ \hline
\end{tabular}
\end{sidewaystable*}

We used an architecture inspired by U-Net~(\cite{ronneberger2015u}), with a downward path of 5 layers of strided convolutions~(\cite{springenberg2014striving}) with 32, 64, 128, 256 and 512 3x3 filters, stride of 2, batch normalization (BN)~(\cite{ioffe2015batch}) and leaky-ReLU activation (LRA)~(\cite{maas2013rectifier}). The upward path consisted of 5 upsampling layers, each one composed of a pair of nearest-neighbor upsampling and a convolutional operation~(\cite{odena2016deconvolution}), with 256, 128, 64, 32 and 3 3x3 filters, BN and LRA; except for the final convolutional layer that did not have BN and used the hyperbolic tangent (tanh) as activation function. We used long skip connections to ease the synthesis upward path~(\cite{ronneberger2015u}), and applied L2 regularization with a factor of \SI{1e-6}{}.

We minimized the mean squared error (MSE) loss using stochastic gradient descent with Adam optimization~(\cite{kingma2014adam}) and 64-sample mini-batch, decreasing the learning rate by a factor of 10 starting from \SI{1e-2} every time the validation loss stopped improving for 4 consecutive epochs until \SI{1e-5}{}. Finally, we selected the weights corresponding to the model with the lowest validation loss during training. 

Convergence to average solutions is a known effect with bottleneck architectures trained with MSE loss. Note, however, that our network-based normalization architecture includes long skip connections between the downward and the upward paths. These skip connections allow the model to copy spatial structures from the input images to the output images with ease, and utilize the rest of the model to modify style-related color features. Since there is no bottleneck effect, i.e., the model has all the information necessary to reconstruct the input image, image reconstructions are highly accurate and do not show any blurriness in practice.

\subsection{Color analysis}

In order to understand how \textit{stain color augmentation} and \textit{stain color normalization} influenced the color differences between internal (\textit{rumc}) and external datasets (rest), we analyzed the image patches in the HSV color space. We measured the mean and standard deviation pixel intensity along the hue and saturation dimensions, and plotted the results in a 2D plane, comparing images processed with the color normalization and augmentation techniques analyzed in this work (see Fig.~\ref{fig:color_space}). We confirmed the clustering effect of normalization algorithms, and the scattering effect of augmentation methods.

\subsection{CNN Classifiers}

In order to measure the effect of \textit{stain color augmentation} and \textit{stain color normalization}, we trained a series of identical CNN classifiers to perform patch classification using different combinations of these techniques. For training and validation purposes, we used the \textit{rumc} datasets described in Sec.~\ref{sec:materials}. 

The architecture of such CNN classifiers consisted of 9 layers of strided convolutions with 32, 64, 64, 128, 128, 256, 256, 512 and 512 3x3 filters, stride of 2 in the even layers, BN and LRA; followed by global average pooling; 50\% dropout; a dense layer with 512 units, BN and LRA; and a linear dense layer with either 2 or 9 units depending on the classification task, followed by a softmax. We applied L2 regularization with a factor of \SI{1e-6}{}.

We minimized the cross-entropy loss using stochastic gradient descent with Adam optimization and 64-sample class-balanced mini-batch, decreasing the learning rate by a factor of 10 starting from \SI{1e-2} every time the validation loss stopped improving for 4 consecutive epochs until \SI{1e-5}{}. Finally, we selected the weights corresponding to the model with the lowest validation loss during training. 

\section{Experimental results}
\label{sec:experiments}

We conducted a series of experiments in order to quantify the impact in performance of the different \textit{stain color augmentation} and \textit{stain color normalization} methods introduced in the previous section across four different classification tasks. We trained a CNN classifier for each combination of organ, color normalization and data augmentation method under consideration. In the case of \textit{grayscale} normalization, we only tested \textit{basic}, \textit{morphology} and \textit{BC} augmentation techniques. We conducted 152 different experiments, repeating each 5 times using different random initialization for the network parameters, accounting for a total of 760 trained CNN classifiers. 

\subsection{Evaluation}
\label{sec:experiments_evaluation}

We evaluated the area under the receiver-operating characteristic curve (AUC) of each CNN in each external test set. In the case of multiclass classification, we considered the unweighted average, i.e. we calculated the individual AUC per label (one-vs-all) and averaged the resulting values. We reported the mean and standard deviation of the resulting AUC for each experiment across five repetitions in Tab.~\ref{tab:ranking}. 

In order to establish a global ranking among methods, shown in the rightmost column in Tab.~\ref{tab:ranking}, we performed the following calculation. We converted the AUC scores into ranking scores per test set column, and averaged these scores along the dataset dimension to obtain a global ranking score per method. Note that we performed an average across ranking scores, rather than AUC scores, following established procedures~(\cite{demvsar2006statistical}). Data in Tab.~\ref{tab:ranking} and the raw AUC scores are provided in machine-readable format as Supplementary Material to this article.

\subsection{Effects of stain color augmentation}

Results in Tab.~\ref{tab:ranking} show that \textit{stain color augmentation} was crucial to obtain top classification performance, regardless of the \textit{stain color normalization} technique used (see top-10 methods). Moreover, note that including color augmentation, either \textit{HSV} or \textit{HED}, was key to obtaining top performance since using \textit{BC} augmentation alone produced mediocre results. We did not find, however, any substantial performance difference between using \textit{HED} or \textit{HSV} color augmentation. Similarly, we found that \textit{strong} and \textit{light} color augmentations achieved similar performance, with a slight advantage towards \textit{light}. Heavy augmentation is known to reduce performance on images similar to those in the training set. However, we found less than 1\% average performance reduction on the internal test set across organs. Regarding non-color augmentation, i.e. \textit{basic}, \textit{morphology} and \textit{BC}, \textit{BC} obtained the best results across almost all \textit{stain color normalization} setups, followed by \textit{morphology} and \textit{basic} augmentation, as expected.

\subsection{Effects of stain color normalization}

According to results in Tab.~\ref{tab:ranking}, overall top performance was achieved without the use of color normalization. This piece of evidence suggests that color normalization is not a necessary condition to achieve high classification performance in histopathology images. However, we observed that color normalization generally produced classifiers that were more robust to different color augmentation techniques, e.g., \textit{Identity} normalization performance diminished with \textit{HSV-light} augmentation whereas \textit{Network} normalization exhibited a high performance regardless of the color augmentation used. 

We did not find any substantial performance difference between neural network based normalization algorithms, \textit{Network} and \textit{Style}. Nevertheless, we observed that none of the classical approaches, \textit{LUT} or \textit{Deconvolution}, surpassed the performance of \textit{Grayscale}. We hypothesize that these classical normalization methods can hide certain useful features from the images, resulting in added input noise that can affect classification performance.

Additionally, we measured the extra time required to normalize a regular whole-slide image composed of $50000\times50000$ RGB pixels. We found \textit{LUT-based} to be the fastest taking \SI{21.8}{min}, followed closely by \textit{network-based} with \SI{26.0}{min}, and the slower \textit{deconv-based} and \textit{style-based} taking \SI{111.2}{min} and \SI{217.8}{min}, respectively, excluding I/O delays.

\section{Discussion}
\label{sec:discussion}

Our experimental results indicate that \textit{stain color augmentation} improved classification performance drastically by increasing the CNN's ability to generalize to unseen stain variations. This was true for most of the experiments regardless of the type of \textit{stain color normalization} technique used. Moreover, we found \textit{HSV} and \textit{HED} color transformations to be the key ingredients to improve performance since removing them, i.e. using \textit{BC} augmentation, yielded a lower AUC under all circumstances; suggesting that inter-lab stain differences were mainly caused by color variations rather than morphological features. Remarkably, we observed hardly any performance difference between \textit{HSV} or \textit{HED}, and \textit{strong} or \textit{light} variation intensity. 

Based on these observations, we concluded that CNNs are mostly insensitive to the type and intensity of the color augmentation used in this setup, as long as one of the methods is used. However, CNNs trained with simpler \textit{stain color normalization} techniques exhibited more sensitivity to the intensity of color augmentation, i.e. they required a stronger augmentation in order to perform well. Finally, the fact that experiments with \textit{grayscale} images achieved mediocre performance was an indication that color provided useful information to the model. The worst performance was achieved with \textit{morphology} and \textit{identity} configurations, which was an indication that color information can act as noise when no augmentation is used, increasing overfitting and generalization error due to stain variation.

Regarding \textit{stain color normalization}, we found that the best performing method did not use any normalization. This result challenged the common assumption that color normalization is a necessary step to achieve top classification performance in the histopathology setting; especially considering that color normalization added a computational overhead that can substantially reduce the overall classification speed. Neural network based methods, both \textit{Network} and \textit{Style}, achieved similar high performance on the benchmark, supporting the idea of reformulating the problem of \textit{stain color normalization} as an image-to-image translation task. 

Furthermore, we observed that all \textit{stain color normalization} techniques obtained a poor performance when no color augmentation was used (below that of \textit{Grayscale} with \textit{BC}). We hypothesize that even in the case of excellent stain normalization, color information can serve as a source of overfitting, worsening with suboptimal normalization. We concluded that using the \textit{stain color normalization} methods evaluated in this paper without proper \textit{stain color augmentation} is insufficient to reduce the generalization error caused by stain variation and results in poor model performance.

Due to computational constraints, we limited the type and number of experiments performed in this study to patch-based classification tasks, ignoring other modalities such as segmentation, instance detection or WSI classification. However, we believe this limitation to have little impact in the conclusions of this study since the problem of generalization error has identical causes and effects in other modalities. In order to reduce the number of experiments, we avoided quantifying the impact of individual augmentation techniques, e.g. scaling augmentation alone, but grouped them into categories instead. Similarly, we limited the hyper-parameters' ranges to certain set of values, e.g. \textit{light} or \textit{strong} stain augmentation intensity. Nevertheless, according to the experimental results, we believe that testing a wider range of hyper-parameter values would not alter the main conclusions of this study.

\section{Conclusion}
\label{sec:conclusion} 
 
For the first time, we quantified the effect of \textit{stain color augmentation} and \textit{stain color normalization} in classification performance across four relevant computational pathology applications using data from 9 different centers. Based on our empirical evaluation, we found that any type of \textit{stain color augmentation}, i.e. \textit{HSV} or \textit{HED} transformation, should always be used. In addition, color augmentation can be combined with neural network based \textit{stain color normalization} to achieve a more robust classification performance. In setups with reduced computational resources, color normalization could be omitted, resulting in a negligible performance reduction and a substantial improvement in processing speed. Finally, we recommend tuning the intensity of the color augmentation to \textit{light} or \textit{strong} in case color normalization is \textit{enabled} or \textit{disabled}, respectively.

\section*{Acknowledgments}

This study was supported by a Junior Researcher grant from the Radboud Institute of Health Sciences (RIHS), Nijmegen, The Netherlands; a grant from the Dutch Cancer Society (KUN 2015-7970); and another grant from the Dutch Cancer Society and the Alpe d'HuZes fund (KUN 2014-7032); this project has also received funding from the European Union's Horizon 2020 research and innovation programme under grant agreement No 825292. The authors would like to thank Dr. Babak Ehteshami Bejnordi for providing the code for the \textit{LUT-based} stain color normalization algorithm; and Canisius-Wilhelmina Ziekenhuis Nijmegen, Laboratorium Pathologie Oost Nederland, University Medical Center Utrecht, and Rijnstate Hospital Arnhem for kindly providing tissue sections for this study.

\bibliographystyle{model2-names}\biboptions{authoryear}
\bibliography{medima-template}

\begin{thebibliography}{37}
\expandafter\ifx\csname natexlab\endcsname\relax\def\natexlab#1{#1}\fi
\providecommand{\url}[1]{\texttt{#1}}
\providecommand{\href}[2]{#2}
\providecommand{\path}[1]{#1}
\providecommand{\DOIprefix}{doi:}
\providecommand{\ArXivprefix}{arXiv:}
\providecommand{\URLprefix}{URL: }
\providecommand{\Pubmedprefix}{pmid:}
\providecommand{\doi}[1]{\href{http://dx.doi.org/#1}{\path{#1}}}
\providecommand{\Pubmed}[1]{\href{pmid:#1}{\path{#1}}}
\providecommand{\bibinfo}[2]{#2}
\ifx\xfnm\relax \def\xfnm[#1]{\unskip,\space#1}\fi
\bibitem[{Albarqouni et~al.(2016)Albarqouni, Baur, Achilles, Belagiannis,
  Demirci and Navab}]{albarqouni2016aggnet}
\bibinfo{author}{Albarqouni, S.}, \bibinfo{author}{Baur, C.},
  \bibinfo{author}{Achilles, F.}, \bibinfo{author}{Belagiannis, V.},
  \bibinfo{author}{Demirci, S.}, \bibinfo{author}{Navab, N.},
  \bibinfo{year}{2016}.
\newblock \bibinfo{title}{Aggnet: deep learning from crowds for mitosis
  detection in breast cancer histology images}.
\newblock \bibinfo{journal}{IEEE Transactions on Medical Imaging}
  \bibinfo{volume}{35}, \bibinfo{pages}{1313--1321}.
\bibitem[{B{\'a}ndi et~al.(2019)B{\'a}ndi, Geessink, Manson, van Dijk,
  Balkenhol, Hermsen, Bejnordi, Lee, Paeng, Zhong et~al.}]{bandi2018detection}
\bibinfo{author}{B{\'a}ndi, P.}, \bibinfo{author}{Geessink, O.},
  \bibinfo{author}{Manson, Q.}, \bibinfo{author}{van Dijk, M.},
  \bibinfo{author}{Balkenhol, M.}, \bibinfo{author}{Hermsen, M.},
  \bibinfo{author}{Bejnordi, B.E.}, \bibinfo{author}{Lee, B.},
  \bibinfo{author}{Paeng, K.}, \bibinfo{author}{Zhong, A.}, et~al.,
  \bibinfo{year}{2019}.
\newblock \bibinfo{title}{{From detection of individual metastases to
  classification of lymph node status at the patient level: the CAMELYON17
  challenge}}.
\newblock \bibinfo{journal}{IEEE Transactions on Medical Imaging}
  \bibinfo{volume}{38}, \bibinfo{pages}{550--560}.
\bibitem[{Bejnordi et~al.(2016)Bejnordi, Litjens, Timofeeva, Otte-H{\"o}ller,
  Homeyer, Karssemeijer and van~der Laak}]{bejnordi2016stain}
\bibinfo{author}{Bejnordi, B.E.}, \bibinfo{author}{Litjens, G.},
  \bibinfo{author}{Timofeeva, N.}, \bibinfo{author}{Otte-H{\"o}ller, I.},
  \bibinfo{author}{Homeyer, A.}, \bibinfo{author}{Karssemeijer, N.},
  \bibinfo{author}{van~der Laak, J.A.}, \bibinfo{year}{2016}.
\newblock \bibinfo{title}{Stain specific standardization of whole-slide
  histopathological images}.
\newblock \bibinfo{journal}{IEEE Transactions on Medical Imaging}
  \bibinfo{volume}{35}, \bibinfo{pages}{404--415}.
\bibitem[{Bejnordi et~al.(2017)Bejnordi, Veta, Van~Diest, Van~Ginneken,
  Karssemeijer, Litjens, Van Der~Laak, Hermsen, Manson, Balkenhol
  et~al.}]{bejnordi2017diagnostic}
\bibinfo{author}{Bejnordi, B.E.}, \bibinfo{author}{Veta, M.},
  \bibinfo{author}{Van~Diest, P.J.}, \bibinfo{author}{Van~Ginneken, B.},
  \bibinfo{author}{Karssemeijer, N.}, \bibinfo{author}{Litjens, G.},
  \bibinfo{author}{Van Der~Laak, J.A.}, \bibinfo{author}{Hermsen, M.},
  \bibinfo{author}{Manson, Q.F.}, \bibinfo{author}{Balkenhol, M.}, et~al.,
  \bibinfo{year}{2017}.
\newblock \bibinfo{title}{Diagnostic assessment of deep learning algorithms for
  detection of lymph node metastases in women with breast cancer}.
\newblock \bibinfo{journal}{JAMA} \bibinfo{volume}{318},
  \bibinfo{pages}{2199--2210}.
\bibitem[{Bug et~al.(2017)Bug, Schneider, Grote, Oswald, Feuerhake, Sch{\"u}ler
  and Merhof}]{bug2017context}
\bibinfo{author}{Bug, D.}, \bibinfo{author}{Schneider, S.},
  \bibinfo{author}{Grote, A.}, \bibinfo{author}{Oswald, E.},
  \bibinfo{author}{Feuerhake, F.}, \bibinfo{author}{Sch{\"u}ler, J.},
  \bibinfo{author}{Merhof, D.}, \bibinfo{year}{2017}.
\newblock \bibinfo{title}{Context-based normalization of histological stains
  using deep convolutional features}, in: \bibinfo{booktitle}{Deep Learning in
  Medical Image Analysis and Multimodal Learning for Clinical Decision
  Support}. \bibinfo{publisher}{Springer}, pp. \bibinfo{pages}{135--142}.
\bibitem[{Bulten et~al.(2019)Bulten, B{\'a}ndi, Hoven, van~de Loo, Lotz, Weiss,
  van~der Laak, van Ginneken, Hulsbergen-van~de Kaa and
  Litjens}]{bulten2019epithelium}
\bibinfo{author}{Bulten, W.}, \bibinfo{author}{B{\'a}ndi, P.},
  \bibinfo{author}{Hoven, J.}, \bibinfo{author}{van~de Loo, R.},
  \bibinfo{author}{Lotz, J.}, \bibinfo{author}{Weiss, N.},
  \bibinfo{author}{van~der Laak, J.}, \bibinfo{author}{van Ginneken, B.},
  \bibinfo{author}{Hulsbergen-van~de Kaa, C.}, \bibinfo{author}{Litjens, G.},
  \bibinfo{year}{2019}.
\newblock \bibinfo{title}{Epithelium segmentation using deep learning in
  {H\&E}-stained prostate specimens with immunohistochemistry as reference
  standard}.
\newblock \bibinfo{journal}{Scientific Reports} \bibinfo{volume}{9},
  \bibinfo{pages}{864}.
\bibitem[{Cho et~al.(2017)Cho, Lim, Choi and Min}]{cho2017neural}
\bibinfo{author}{Cho, H.}, \bibinfo{author}{Lim, S.}, \bibinfo{author}{Choi,
  G.}, \bibinfo{author}{Min, H.}, \bibinfo{year}{2017}.
\newblock \bibinfo{title}{Neural stain-style transfer learning using {GAN} for
  histopathological images}, in: \bibinfo{booktitle}{Asian Conference on
  Machine Learning}.
\bibitem[{Ciompi et~al.(2017)Ciompi, Geessink, Bejnordi, de~Souza, Baidoshvili,
  Litjens, van Ginneken, Nagtegaal and van~der Laak}]{ciompi2017importance}
\bibinfo{author}{Ciompi, F.}, \bibinfo{author}{Geessink, O.},
  \bibinfo{author}{Bejnordi, B.E.}, \bibinfo{author}{de~Souza, G.S.},
  \bibinfo{author}{Baidoshvili, A.}, \bibinfo{author}{Litjens, G.},
  \bibinfo{author}{van Ginneken, B.}, \bibinfo{author}{Nagtegaal, I.},
  \bibinfo{author}{van~der Laak, J.}, \bibinfo{year}{2017}.
\newblock \bibinfo{title}{The importance of stain normalization in colorectal
  tissue classification with convolutional networks}, in:
  \bibinfo{booktitle}{Biomedical Imaging (ISBI 2017), 2017 IEEE 14th
  International Symposium on}, \bibinfo{organization}{IEEE}. pp.
  \bibinfo{pages}{160--163}.
\bibitem[{Clarke and Treanor(2017)}]{clarke2017colour}
\bibinfo{author}{Clarke, E.L.}, \bibinfo{author}{Treanor, D.},
  \bibinfo{year}{2017}.
\newblock \bibinfo{title}{Colour in digital pathology: a review}.
\newblock \bibinfo{journal}{Histopathology} \bibinfo{volume}{70},
  \bibinfo{pages}{153--163}.
\bibitem[{Dem{\v{s}}ar(2006)}]{demvsar2006statistical}
\bibinfo{author}{Dem{\v{s}}ar, J.}, \bibinfo{year}{2006}.
\newblock \bibinfo{title}{Statistical comparisons of classifiers over multiple
  data sets}.
\newblock \bibinfo{journal}{Journal of Machine Learning Research}
  \bibinfo{volume}{7}, \bibinfo{pages}{1--30}.
\bibitem[{Deng et~al.(2009)Deng, Dong, Socher, Li, Li and
  Fei-Fei}]{deng2009imagenet}
\bibinfo{author}{Deng, J.}, \bibinfo{author}{Dong, W.},
  \bibinfo{author}{Socher, R.}, \bibinfo{author}{Li, L.J.},
  \bibinfo{author}{Li, K.}, \bibinfo{author}{Fei-Fei, L.},
  \bibinfo{year}{2009}.
\newblock \bibinfo{title}{Imagenet: A large-scale hierarchical image database},
  in: \bibinfo{booktitle}{Computer Vision and Pattern Recognition, 2009. CVPR
  2009. IEEE Conference on}, \bibinfo{organization}{IEEE}. pp.
  \bibinfo{pages}{248--255}.
\bibitem[{Gertych et~al.(2015)Gertych, Ing, Ma, Fuchs, Salman, Mohanty, Bhele,
  Vel{\'a}squez-Vacca, Amin and Knudsen}]{gertych2015machine}
\bibinfo{author}{Gertych, A.}, \bibinfo{author}{Ing, N.}, \bibinfo{author}{Ma,
  Z.}, \bibinfo{author}{Fuchs, T.J.}, \bibinfo{author}{Salman, S.},
  \bibinfo{author}{Mohanty, S.}, \bibinfo{author}{Bhele, S.},
  \bibinfo{author}{Vel{\'a}squez-Vacca, A.}, \bibinfo{author}{Amin, M.B.},
  \bibinfo{author}{Knudsen, B.S.}, \bibinfo{year}{2015}.
\newblock \bibinfo{title}{Machine learning approaches to analyze histological
  images of tissues from radical prostatectomies}.
\newblock \bibinfo{journal}{Computerized Medical Imaging and Graphics}
  \bibinfo{volume}{46}, \bibinfo{pages}{197--208}.
\bibitem[{Goodfellow et~al.(2014)Goodfellow, Pouget-Abadie, Mirza, Xu,
  Warde-Farley, Ozair, Courville and Bengio}]{goodfellow2014generative}
\bibinfo{author}{Goodfellow, I.}, \bibinfo{author}{Pouget-Abadie, J.},
  \bibinfo{author}{Mirza, M.}, \bibinfo{author}{Xu, B.},
  \bibinfo{author}{Warde-Farley, D.}, \bibinfo{author}{Ozair, S.},
  \bibinfo{author}{Courville, A.}, \bibinfo{author}{Bengio, Y.},
  \bibinfo{year}{2014}.
\newblock \bibinfo{title}{Generative adversarial nets}, in:
  \bibinfo{booktitle}{Advances in Neural Information Processing Systems}, pp.
  \bibinfo{pages}{2672--2680}.
\bibitem[{Goodfellow et~al.(2016)}]{Goodfellow-et-al-2016}
\bibinfo{author}{Goodfellow, I.}, et~al., \bibinfo{year}{2016}.
\newblock \bibinfo{title}{Deep Learning}.
\newblock \bibinfo{publisher}{MIT Press}.
\newblock \bibinfo{note}{\url{http://www.deeplearningbook.org}}.
\bibitem[{Haeberli and Voorhies(1994)}]{haeberli1994image}
\bibinfo{author}{Haeberli, P.}, \bibinfo{author}{Voorhies, D.},
  \bibinfo{year}{1994}.
\newblock \bibinfo{title}{Image processing by linear interpolation and
  extrapolation}.
\newblock \bibinfo{journal}{IRIS Universe Magazine} \bibinfo{volume}{28},
  \bibinfo{pages}{8--9}.
\bibitem[{Ioffe and Szegedy(2015)}]{ioffe2015batch}
\bibinfo{author}{Ioffe, S.}, \bibinfo{author}{Szegedy, C.},
  \bibinfo{year}{2015}.
\newblock \bibinfo{title}{Batch normalization: Accelerating deep network
  training by reducing internal covariate shift}, in:
  \bibinfo{booktitle}{International Conference on Machine Learning}, pp.
  \bibinfo{pages}{448--456}.
\bibitem[{Janowczyk et~al.(2017)Janowczyk, Basavanhally and
  Madabhushi}]{janowczyk2017stain}
\bibinfo{author}{Janowczyk, A.}, \bibinfo{author}{Basavanhally, A.},
  \bibinfo{author}{Madabhushi, A.}, \bibinfo{year}{2017}.
\newblock \bibinfo{title}{Stain normalization using sparse autoencoders
  (stanosa): application to digital pathology}.
\newblock \bibinfo{journal}{Computerized Medical Imaging and Graphics}
  \bibinfo{volume}{57}, \bibinfo{pages}{50--61}.
\bibitem[{Kather et~al.(2016)Kather, Weis, Bianconi, Melchers, Schad, Gaiser,
  Marx and Z{\"o}llner}]{kather2016multi}
\bibinfo{author}{Kather, J.N.}, \bibinfo{author}{Weis, C.A.},
  \bibinfo{author}{Bianconi, F.}, \bibinfo{author}{Melchers, S.M.},
  \bibinfo{author}{Schad, L.R.}, \bibinfo{author}{Gaiser, T.},
  \bibinfo{author}{Marx, A.}, \bibinfo{author}{Z{\"o}llner, F.G.},
  \bibinfo{year}{2016}.
\newblock \bibinfo{title}{Multi-class texture analysis in colorectal cancer
  histology}.
\newblock \bibinfo{journal}{Scientific Reports} \bibinfo{volume}{6},
  \bibinfo{pages}{27988}.
\bibitem[{Khan et~al.(2014)}]{khan2014nonlinear}
\bibinfo{author}{Khan, A.M.}, et~al., \bibinfo{year}{2014}.
\newblock \bibinfo{title}{A nonlinear mapping approach to stain normalization
  in digital histopathology images using image-specific color deconvolution}.
\newblock \bibinfo{journal}{IEEE Transactions on Biomedical Engineering}
  \bibinfo{volume}{61}, \bibinfo{pages}{1729--1738}.
\bibitem[{Kingma and Ba(2014)}]{kingma2014adam}
\bibinfo{author}{Kingma, D.P.}, \bibinfo{author}{Ba, J.}, \bibinfo{year}{2014}.
\newblock \bibinfo{title}{Adam: A method for stochastic optimization}, in:
  \bibinfo{booktitle}{International Conference on Learning Representations}.
\bibitem[{Kingma and Welling(2013)}]{kingma2013auto}
\bibinfo{author}{Kingma, D.P.}, \bibinfo{author}{Welling, M.},
  \bibinfo{year}{2013}.
\newblock \bibinfo{title}{Auto-encoding variational bayes}, in:
  \bibinfo{booktitle}{International Conference on Learning Representations}.
\bibitem[{Komura and Ishikawa(2018)}]{komura2018machine}
\bibinfo{author}{Komura, D.}, \bibinfo{author}{Ishikawa, S.},
  \bibinfo{year}{2018}.
\newblock \bibinfo{title}{Machine learning methods for histopathological image
  analysis}.
\newblock \bibinfo{journal}{Computational and structural biotechnology journal}
  \bibinfo{volume}{16}, \bibinfo{pages}{34--42}.
\bibitem[{Liu et~al.(2017)Liu, Gadepalli, Norouzi, Dahl, Kohlberger, Boyko,
  Venugopalan, Timofeev, Nelson, Corrado et~al.}]{liu2017detecting}
\bibinfo{author}{Liu, Y.}, \bibinfo{author}{Gadepalli, K.},
  \bibinfo{author}{Norouzi, M.}, \bibinfo{author}{Dahl, G.E.},
  \bibinfo{author}{Kohlberger, T.}, \bibinfo{author}{Boyko, A.},
  \bibinfo{author}{Venugopalan, S.}, \bibinfo{author}{Timofeev, A.},
  \bibinfo{author}{Nelson, P.Q.}, \bibinfo{author}{Corrado, G.S.}, et~al.,
  \bibinfo{year}{2017}.
\newblock \bibinfo{title}{Detecting cancer metastases on gigapixel pathology
  images}.
\newblock \bibinfo{journal}{arXiv preprint arXiv:1703.02442} .
\bibitem[{Maas et~al.(2013)Maas, Hannun and Ng}]{maas2013rectifier}
\bibinfo{author}{Maas, A.L.}, \bibinfo{author}{Hannun, A.Y.},
  \bibinfo{author}{Ng, A.Y.}, \bibinfo{year}{2013}.
\newblock \bibinfo{title}{Rectifier nonlinearities improve neural network
  acoustic models}, in: \bibinfo{booktitle}{International Conference on Machine
  Learning}.
\bibitem[{Macenko et~al.(2009)}]{macenko2009method}
\bibinfo{author}{Macenko, M.}, et~al., \bibinfo{year}{2009}.
\newblock \bibinfo{title}{A method for normalizing histology slides for
  quantitative analysis}, in: \bibinfo{booktitle}{Biomedical Imaging: From Nano
  to Macro, 2009. ISBI'09. IEEE International Symposium on},
  \bibinfo{organization}{IEEE}. pp. \bibinfo{pages}{1107--1110}.
\bibitem[{Odena et~al.(2016)Odena, Dumoulin and Olah}]{odena2016deconvolution}
\bibinfo{author}{Odena, A.}, \bibinfo{author}{Dumoulin, V.},
  \bibinfo{author}{Olah, C.}, \bibinfo{year}{2016}.
\newblock \bibinfo{title}{Deconvolution and checkerboard artifacts}.
\newblock \bibinfo{journal}{Distill} \bibinfo{volume}{1}.
\bibitem[{Reinhard et~al.(2001)Reinhard, Adhikhmin, Gooch and
  Shirley}]{reinhard2001color}
\bibinfo{author}{Reinhard, E.}, \bibinfo{author}{Adhikhmin, M.},
  \bibinfo{author}{Gooch, B.}, \bibinfo{author}{Shirley, P.},
  \bibinfo{year}{2001}.
\newblock \bibinfo{title}{Color transfer between images}.
\newblock \bibinfo{journal}{IEEE Computer Graphics and Applications}
  \bibinfo{volume}{21}, \bibinfo{pages}{34--41}.
\bibitem[{Ronneberger et~al.(2015)Ronneberger, Fischer and
  Brox}]{ronneberger2015u}
\bibinfo{author}{Ronneberger, O.}, \bibinfo{author}{Fischer, P.},
  \bibinfo{author}{Brox, T.}, \bibinfo{year}{2015}.
\newblock \bibinfo{title}{U-net: Convolutional networks for biomedical image
  segmentation}, in: \bibinfo{booktitle}{International Conference on Medical
  image computing and computer-assisted intervention},
  \bibinfo{organization}{Springer}. pp. \bibinfo{pages}{234--241}.
\bibitem[{Simard et~al.(2003)Simard, Steinkraus, Platt et~al.}]{simard2003best}
\bibinfo{author}{Simard, P.Y.}, \bibinfo{author}{Steinkraus, D.},
  \bibinfo{author}{Platt, J.C.}, et~al., \bibinfo{year}{2003}.
\newblock \bibinfo{title}{Best practices for convolutional neural networks
  applied to visual document analysis.}, in: \bibinfo{booktitle}{International
  Conference on Document Analysis and Recognition},
  \bibinfo{organization}{IEEE}. pp. \bibinfo{pages}{958--962}.
\bibitem[{Sirinukunwattana et~al.(2017)Sirinukunwattana, Pluim, Chen, Qi, Heng,
  Guo, Wang, Matuszewski, Bruni, Sanchez et~al.}]{sirinukunwattana2017gland}
\bibinfo{author}{Sirinukunwattana, K.}, \bibinfo{author}{Pluim, J.P.},
  \bibinfo{author}{Chen, H.}, \bibinfo{author}{Qi, X.}, \bibinfo{author}{Heng,
  P.A.}, \bibinfo{author}{Guo, Y.B.}, \bibinfo{author}{Wang, L.Y.},
  \bibinfo{author}{Matuszewski, B.J.}, \bibinfo{author}{Bruni, E.},
  \bibinfo{author}{Sanchez, U.}, et~al., \bibinfo{year}{2017}.
\newblock \bibinfo{title}{Gland segmentation in colon histology images: The
  glas challenge contest}.
\newblock \bibinfo{journal}{Medical Image Analysis} \bibinfo{volume}{35},
  \bibinfo{pages}{489--502}.
\bibitem[{Springenberg et~al.(2014)Springenberg, Dosovitskiy, Brox and
  Riedmiller}]{springenberg2014striving}
\bibinfo{author}{Springenberg, J.T.}, \bibinfo{author}{Dosovitskiy, A.},
  \bibinfo{author}{Brox, T.}, \bibinfo{author}{Riedmiller, M.},
  \bibinfo{year}{2014}.
\newblock \bibinfo{title}{Striving for simplicity: The all convolutional net},
  in: \bibinfo{booktitle}{International Conference on Learning
  Representations}.
\bibitem[{Tellez et~al.(2018)Tellez, Balkenhol, Otte-H{\"o}ller, van~de Loo,
  Vogels, Bult, Wauters, Vreuls, Mol, Karssemeijer et~al.}]{tellez2018whole}
\bibinfo{author}{Tellez, D.}, \bibinfo{author}{Balkenhol, M.},
  \bibinfo{author}{Otte-H{\"o}ller, I.}, \bibinfo{author}{van~de Loo, R.},
  \bibinfo{author}{Vogels, R.}, \bibinfo{author}{Bult, P.},
  \bibinfo{author}{Wauters, C.}, \bibinfo{author}{Vreuls, W.},
  \bibinfo{author}{Mol, S.}, \bibinfo{author}{Karssemeijer, N.}, et~al.,
  \bibinfo{year}{2018}.
\newblock \bibinfo{title}{{Whole-Slide Mitosis Detection in H\&E Breast
  Histology Using PHH3 as a Reference to Train Distilled Stain-Invariant
  Convolutional Networks}}.
\newblock \bibinfo{journal}{IEEE Transactions on Medical Imaging}
  \bibinfo{volume}{37}, \bibinfo{pages}{2126--2136}.
\bibitem[{Veta et~al.(2019)Veta, Heng, Stathonikos, Bejnordi, Beca, Wollmann,
  Rohr, Shah, Wang, Rousson et~al.}]{veta2018predicting}
\bibinfo{author}{Veta, M.}, \bibinfo{author}{Heng, Y.J.},
  \bibinfo{author}{Stathonikos, N.}, \bibinfo{author}{Bejnordi, B.E.},
  \bibinfo{author}{Beca, F.}, \bibinfo{author}{Wollmann, T.},
  \bibinfo{author}{Rohr, K.}, \bibinfo{author}{Shah, M.A.},
  \bibinfo{author}{Wang, D.}, \bibinfo{author}{Rousson, M.}, et~al.,
  \bibinfo{year}{2019}.
\newblock \bibinfo{title}{Predicting breast tumor proliferation from
  whole-slide images: the {TUPAC16} challenge}.
\newblock \bibinfo{journal}{Medical Image Analysis} \bibinfo{volume}{54},
  \bibinfo{pages}{111--121}.
\bibitem[{Van~der Walt et~al.(2014)Van~der Walt, Sch{\"o}nberger,
  Nunez-Iglesias, Boulogne, Warner, Yager, Gouillart and Yu}]{van2014scikit}
\bibinfo{author}{Van~der Walt, S.}, \bibinfo{author}{Sch{\"o}nberger, J.L.},
  \bibinfo{author}{Nunez-Iglesias, J.}, \bibinfo{author}{Boulogne, F.},
  \bibinfo{author}{Warner, J.D.}, \bibinfo{author}{Yager, N.},
  \bibinfo{author}{Gouillart, E.}, \bibinfo{author}{Yu, T.},
  \bibinfo{year}{2014}.
\newblock \bibinfo{title}{Scikit-image: image processing in python}.
\newblock \bibinfo{journal}{PeerJ} \bibinfo{volume}{2}.
\bibitem[{Wang et~al.(2015)Wang, Foran, Ren, Zhong, Kim and
  Qi}]{wang2015exploring}
\bibinfo{author}{Wang, D.}, \bibinfo{author}{Foran, D.J.},
  \bibinfo{author}{Ren, J.}, \bibinfo{author}{Zhong, H.}, \bibinfo{author}{Kim,
  I.Y.}, \bibinfo{author}{Qi, X.}, \bibinfo{year}{2015}.
\newblock \bibinfo{title}{Exploring automatic prostate histopathology image
  gleason grading via local structure modeling}, in: \bibinfo{booktitle}{2015
  37th Annual International Conference of the IEEE Engineering in Medicine and
  Biology Society (EMBC)}, \bibinfo{organization}{IEEE}. pp.
  \bibinfo{pages}{2649--2652}.
\bibitem[{Zanjani et~al.(2018)Zanjani, Zinger, Bejnordi, van~der Laak and
  de~With}]{zanjani2018stain}
\bibinfo{author}{Zanjani, F.G.}, \bibinfo{author}{Zinger, S.},
  \bibinfo{author}{Bejnordi, B.E.}, \bibinfo{author}{van~der Laak, J.A.},
  \bibinfo{author}{de~With, P.H.}, \bibinfo{year}{2018}.
\newblock \bibinfo{title}{Stain normalization of histopathology images using
  generative adversarial networks}, in: \bibinfo{booktitle}{International
  Symposium on Biomedical Imaging}, \bibinfo{organization}{IEEE}. pp.
  \bibinfo{pages}{573--577}.
\bibitem[{Zhu et~al.(2014)Zhu, Zhang, Liu and Metaxas}]{zhu2014scalable}
\bibinfo{author}{Zhu, Y.}, \bibinfo{author}{Zhang, S.}, \bibinfo{author}{Liu,
  W.}, \bibinfo{author}{Metaxas, D.N.}, \bibinfo{year}{2014}.
\newblock \bibinfo{title}{Scalable histopathological image analysis via active
  learning}, in: \bibinfo{booktitle}{International Conference on Medical Image
  Computing and Computer-Assisted Intervention},
  \bibinfo{organization}{Springer}. pp. \bibinfo{pages}{369--376}.

\end{thebibliography}


\end{document}